\documentclass[journal]{IEEEtran} 

\ifCLASSOPTIONcompsoc
\usepackage[nocompress]{cite}
\else
\usepackage{cite}
\fi
\ifCLASSINFOpdf
\usepackage[pdftex]{graphicx}
\DeclareGraphicsExtensions{.pdf,.jpeg,.png}
\else
\fi
\usepackage{times}
\usepackage{helvet}
\usepackage{courier}
\usepackage{amsmath,amsfonts}
\usepackage{algorithmic}
\usepackage{algorithm}  
\usepackage{array}
\usepackage{textcomp}
\usepackage{url}
\usepackage{verbatim}
\usepackage{graphicx}
\usepackage{ntheorem}
\usepackage{enumerate}
\usepackage{subfigure}
\usepackage{booktabs}
\usepackage{bbding}
\usepackage{multirow}
\usepackage{multicol}
\usepackage[caption=false,font=footnotesize,labelfont=rm,textfont=rm]{subfig}
\usepackage{amssymb}
 
\usepackage{xcolor}

\begin{document}
%
\title{TERM Model: Tensor Ring Mixture Model for Density Estimation}
\author{Ruituo Wu, Jiani Liu, Ce Zhu, \emph{Fellow}, \emph{IEEE}, Anh-Huy Phan,  \emph{Member}, \emph{IEEE}, Ivan V. Oseledets, and Yipeng Liu \emph{Senior Member}, \emph{IEEE}. \thanks{This work was supported in part by the National Natural Science Foundation of China (NSFC) under Grant 62171088, Grant 6220106011, and Grant U19A2052. Yipeng Liu is the corresponding author.}

       \thanks{R. Wu, J. Liu C. Zhu and Y. Liu are with the School of Information and Communication Engineering, University of Electronic Science and Technology of China (UESTC), Chengdu, 611731, China. E-mail: yipengliu@uestc.edu.cn}.
        \thanks{A. Phan and I. Oseledets are with the Skolkovo Institute of Science and Technology (Skoltech), Moscow 143026, Russia.}}
\maketitle

\begin{abstract}

Efficient probability density estimation is a core challenge in statistical machine learning. Tensor-based probabilistic graph methods address interpretability and stability concerns encountered in neural network approaches. 
However, a substantial number of potential tensor permutations can lead to a tensor network with the same structure but varying expressive capabilities.
In this paper, we take tensor ring decomposition for density estimator, which significantly reduces the number of permutation candidates while enhancing expressive capability compared with existing used decompositions.
Additionally, a mixture model that incorporates multiple permutation candidates with adaptive weights is further designed, resulting in increased expressive flexibility and comprehensiveness. Different from the prevailing directions of tensor network structure/permutation search, our approach provides a new viewpoint inspired by ensemble learning. This approach acknowledges that suboptimal permutations can offer distinctive information besides that of optimal permutations. Experiments show the superiority of the proposed approach in estimating probability density for moderately dimensional datasets and sampling to capture intricate details.

\end{abstract}

\section{Introduction}
Density estimation approximates the probability density function under the assumption of independent and uniformly distributed data. Density estimation plays a crucial role in applications like anomaly detection \cite{nachman2020anomaly}, outlier detection \cite{latecki2007outlier}, clustering \cite{campello2013cluster}, classification \cite{oyang2005classification}, data generation \cite{papamakarios2017masked}, and data compression \cite{alsing2018compress}. 

Early non-parametric methods, such as histogram estimation \cite{van1973histogram} and kernel density estimation \cite{terrell1992variable}, often show poor performance when dealing with even moderate dimensional data. Recently artificial neural network-based approaches address this limitation, including the variational autoencoder (VAE) \cite{kingma2013auto}, generative adversarial networks (GAN) \cite{goodfellow2014generative}, auto-regressive neural networks \cite{van2016pixel}, and invertible flows \cite{dinh2016density}. While these neural network-based methods have shown promise, they often require careful tuning of hyperparameters and architectures yet may lack interpretability. For instance, GANs \cite{goodfellow2014generative} do not have a tractable log-likelihood for training, whereas VAEs \cite{kingma2013auto} use the evidence lower bound (ELBO) as a substitute. Modern flow-based generative models like Real NVP \cite{dinh2016density}, Glow \cite{kingma2018glow}, and FFJORD  \cite{grathwohl2018ffjord} employ unbiased estimates for training instead of optimizing the true log-likelihood. Furthermore, the aforementioned model is unable to simulate both marginal distributions and conditional probability densities.

In recent years, tensor decomposition-based methods have shown promise in various tasks \cite{liu2022tensor}, such as completion \cite{huang2020robust} and regression \cite{liu2020smooth,liu2021tensor}. In the realm of probability density estimation, tensor-based method efficiently circumvents the challenges posed by unstable adversarial training and parameter tuning encountered in neural network-based approaches. Previous research mainly focuses on two tensor decomposition approaches, including canonical polyadic decomposition (CPD) \cite{hitchcock1927expression} and tensor train (TT) decomposition \cite{oseledets2011tensor}. However, determining the optimal rank for CPD remains
an NP-hard problem \cite{hitchcock1927expression}, while TT is organized based on a chain of variables. 
Although the TT format offers advantages in terms of computational and storage costs, its chain structure limits its expressive capability. Specifically, an imbalanced distribution of TT ranks can lead to a potential loss of accuracy when approximating tensors \cite{zhao2016tensor}. This effect is particularly pronounced when data information is concentrated near the boundaries of the tensor.
In addition, several studies reveal that mapping tensor modes onto the tensor network (TN) vertices significantly impacts the model expressive power \cite{ye2018tensor,li2020high,qiu2021memory}. Different `mode-vertex' mapping has different expressive and generalization abilities \cite{li2022permutation}. Driven by this, tensor network structure search (TN-SS) \cite{li2020evolutionary} and tensor network permutation search (TN-PS) \cite{li2022permutation} approaches are developed to discover optimal structures or permutations based on the data itself. 

However, we posit that while optimal permutations are suitable for the current data, other permutations might contain valuable information, particularly in non-parametric probability density estimation tasks where prior knowledge about data dimension dependencies is lacking. For example, in Gaussian mixture models \cite{reynolds2009gaussian}, even the Gaussian components with smaller weights contribute meaningfully to the final density assessment. Furthermore, we believe that searching for the optimal tensor structure using the exhaustive method on mid-dimensional data is nearly impossible due to the exponential increase in the number of tensor structure candidates, as discussed in \cite{li2022permutation}.

These challenges prompt us to explore the use of tensor ring decomposition (TR) \cite{zhao2016tensor} for probability density estimation. TR provides a powerful expressive capability and more balanced ranks compared with TT \cite{long2021bayesian}. Furthermore, TR retains the benefits of linear storage and computing cost efficiency. TR's unique rotational invariance results in its number of decomposition permutations being analogous to an undirected circular arrangement with $\frac{(D-1)!}{2}$ permutations, which is $D$ times fewer than TT's linear arrangement with $\frac{D!}{2}$ permutations\cite{li2022permutation}. Here, $D$ represents the data dimension.

\begin{figure}[!tbp]
        \centering
		{\includegraphics[scale=0.5]{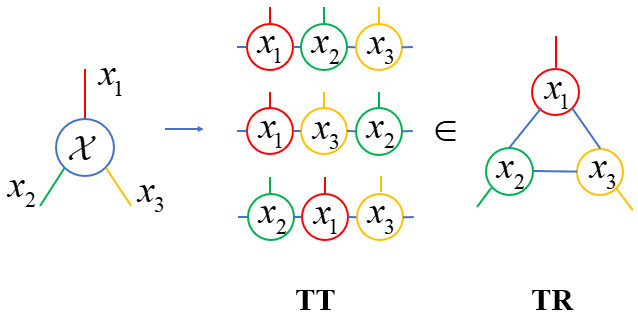}}
		\caption{The TT and TR decompositions of a third-order tensor under various permutations.} 
        \label{TTTR}
\end{figure}

For example, for a third-order tensor $\mathcal{X}$, the number of permutation candidates is $3$ for TT and $1$ for TR, as depicted in Fig.\ref{TTTR}.
The TT permutations can be viewed as a special case, in the undirected circular permutation of TR.
In light of these considerations, we draw inspiration from ensemble learning and classical Gaussian mixture models, proposing a novel non-parametric density estimator. We first design a `basis learner' by expanding the density function in some uniform B-spline basis with coefficients approximated by TR factors. Subsequently, a mixture model is constructed by amalgamating basis learners built on different permutation candidates. These learners are allocated adaptive weights to augment the model's flexibility and adaptability. The main contributions of this paper can be summarized as follows.
\begin{itemize}
    \item  We propose a novel density estimation method called tensor ring density estimator (TRDE), which approximates the coefficients for the expansion of the density function in some uniform B-spline basis by TR factorization. It allows exact sampling and efficient computation of the cumulative probability density function, marginal probability density function, and its derivatives.

    \item Based on the basic learner TRDE, we introduce a mixture model TERM, which ensembles various permutation structures and adaptively learns the weight of each basis learner. This is the initial attempt at implementing a mixture model based on TR. This approach significantly reduces the number of permutation candidates and enables a smaller number of components to comprehensively represent the underlying structural information. Furthermore, the simultaneous sampling technique adds minimal computational overhead.

    \item 
    Experiments show the superiority of the proposed approach in estimating probability density for moderately dimensional datasets and sampling to capture intricate details.  
    
\end{itemize}

\section{Related work}
The roots of non-parametric probability density estimation can be traced back to the early methods of histogram estimation \cite{van1973histogram} and kernel density estimation \cite{terrell1992variable}. While these methods are valuable in lower-dimensional settings, they often struggle to cope with the challenges of moderate dimensions. 

In recent years, the remarkable progress in neural networks and deep learning has given rise to novel models and techniques for probability density estimation. One prominent example is VAE \cite{kingma2013auto}, which aims to minimize the divergence between the data and the prior distribution by optimizing the ELBO. Another notable approach is GAN \cite{goodfellow2014generative}, which employs adversarial learning between a generator and a discriminator to generate highly realistic samples. While VAEs and GANs excel at generating realistic images, their practical applications in statistical environments are constrained by the intractable partition functions they involve. As a result, these models may struggle to perform well in scenarios that demand precise probability density estimation and statistical analysis. 

The energy-based models \cite{lecun2006tutorial} aim to maximize the logarithmic likelihood and approximate the partition function using Markov Chain Monte Carlo (MCMC) sampling. However, due to the reliance on MCMC sampling, these models can only provide approximate solutions. This limitation affects the quality and efficiency of sampling from these models. Autoregressive models \cite{van2016pixel}, on the other hand, assume a specific feature ordering and factorize the joint distribution into a product of conditional probabilities. Consequently, the permutation of features becomes a crucial factor in such models. Moreover, sampling from autoregressive models can be computationally expensive, making them less practical for certain applications.

The family of normalizing flows \cite{papamakarios2021normalizing} offers a flexible framework for transforming simple probability distributions, like multivariate Gaussians, into more complex distributions capable of capturing the data distribution. However, finding suitable invertible transformations and computing higher-order Jacobian matrices can be computationally challenging. Furthermore, normalizing flows lack the ability to calculate marginal probability density, cumulative probability density, and their derivatives accurately. 

Recently, tensor-based models have gained considerable attention in the field of probability density estimation. One notable approach is CPDE \cite{amiridi2022low}, which leverages CPD and truncates the high-dimensional Fourier transform twice to reduce computational complexity. However, determining the optimal rank for CPD remains an NP-hard problem\cite{kolda2009tensor}, presenting a challenge in achieving the most efficient and accurate results.

In contrast, a series of works based on the TT format \cite{dolgov2020approximation,cui2022deep,novikov2021tensor} addresses several challenges by providing explicit density functions, computable partition functions, highly interpretable models and parameters, as well as linear storage costs and computational efficiency. However, the linear structure of the TT decomposition highlights the importance of selecting an appropriate feature permutation. The reliance on a single probability graph structure may limit the model's expressive capacity, potentially leading to sub-optimal performance in capturing complex data distributions.

\section{Notations and Preliminaries}
\subsection{Notations}

Throughout the paper, we use different notations to represent scalar, vector, matrix, and tensor variables, i.e., a scalar as $x$, a vector as $\mathbf{x}$, a matrix as $\mathbf{X}$, and a tensor as $\mathcal{X}$. The trace operation of a square matrix $\mathbf{X}$ is denoted as $\operatorname{Trace}(\mathbf{X}) = \sum_{i=1}^I x_{i,i}$, where $x_{i,i}$ represents the $i$-th diagonal element.
A tensor of order $D$ is represented by $\mathcal{X} \in \mathbb{R}^{I_1 \times I_2 \times \cdots \times I_D}$, whose $i_1,\ldots,i_D$-th element is denoted as $\mathcal{X}(i_1,\ldots,i_D)$, $i_d$ for $d \in {1, \ldots, D}$ are the indices along each dimension of the tensor. 
The inner product between two tensors, $\mathcal{X}, \mathcal{Y}$ $\in \mathbb{R}^{I_1 \times I_2 \times \ldots \times I_D}$ yields a scalar computed as 
\begin{equation}
\begin{aligned}
\langle \mathcal{X}, \mathcal{Y} \rangle = \sum_{i_1=1}^{I_1} \cdots \sum_{i_D=1}^{I_D} \mathcal{X}(i_1, \dots, i_D) \mathcal{Y}(i_1, \dots, i_D).
\end{aligned}
\end{equation}

The tensor outer product between two tensors, $\mathcal{X}\in \mathbb{R}^{I_1  \times \cdots \times I_D}$, $\mathcal{Y}\in \mathbb{R}^{K_1 \times  \cdots \times K_M}$,  yields a tensor $\mathcal{Z}\in \mathbb{R}^{I_1  \times \cdots \times I_D\times K_1 \times \ldots \times K_M}$, whose elements are computed as
\begin{equation}
\mathcal{Z}(i_1,\ldots, i_D,k_1,\ldots,k_M)
=\mathcal{X}({i_1, \ldots, i_D})  \mathcal{Y}({k_1, \ldots, k_M}).
\end{equation}
We can write the tensor outer product as $\mathcal{Z} =\mathcal{X}\circ \mathcal{Y} $.

Einstein summation notation is expressed as $\operatorname{einsum}$, which is commonly used in tensor calculus and linear algebra. This notation allows us to compactly represent and perform calculations involving tensors and multi-dimensional arrays. In Einstein summation notation, repeated indices in a term imply summation over all possible values of that index. For example, given two matrices $\mathbf{A}\in \mathbb{R}^{I \times K} $ and $\mathbf{B}\in \mathbb{R}^{K \times J}$, matrix product of these two matrices $\sum_{k=1}^K A_{i k} B_{k j}$ can be expressed as 
$res \leftarrow \operatorname{einsum}({ }^{\prime} i k, kj \rightarrow i  j ^{\prime},\mathbf{A},\mathbf{B})$. Here, $res$ represents the result of the operation.

\subsection{Tensor Ring  Decomposition}
\begin{figure}[htbp!]
        \centering
		\subfigure{\includegraphics[scale=0.6]{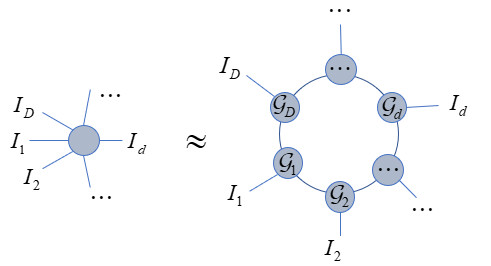}}
	\caption{TR decomposition} 
        \label{TR}
	\end{figure}

Tensor ring decomposition is a tensor factorization technique used to represent high-dimensional tensors as a series of lower-dimensional tensors. Given a tensor $\mathcal{X} \in \mathbb{R}^{I_1 \times \cdots \times I_D}$, the TR decomposition is defined as follows:
\begin{equation}
\begin{aligned}
& \mathcal{X}\left(i_1, i_2, \ldots, i_D\right) \\
&= \operatorname{Trace}\left(\mathcal{G}_1\left(:, i_1, :\right) \mathcal{G}_2\left(:, i_2, :\right) \ldots \mathcal{G}_D\left(:, i_D, :\right)\right),
\end{aligned}
\end{equation}
where $\mathcal{G}_d \in \mathbb{R}^{R_{d-1} \times I_d \times R_{d}}$ is the $d$-th core tensor, and TR ranks are defined as ${R_0, \ldots, R_D}$ with $R_D = R_0$. The graphical illustration of TR decomposition is shown in Fig.\ref{TR} 
 
 In the  rest of this paper,  we use $\{{\mathcal{G}}_d^{(\mathcal{X})}\}_{d=1}^D$ to represent the TR cores of a tensor. 
And the TR rank of each TR core tensor is equal without special emphasis. Lateral slices refer to two-dimensional sections of a tensor, defined by holding all but two rank indices constant. In the case of a third-order TR core tensor $\mathcal{G} \in \mathbb{R}^{R_{d-1} \times I_d \times R_d}$, a lateral slice is represented as $\mathcal{G}\left(:, i_d,:\right)\in \mathbb{R}^{R_{d-1} \times R_d} $. 
\subsection{Tensor Inner product calculation for TR format }
Due to the storage overhead in higher-order tensors, performing an inner product operation on the entire tensor is often infeasible. Here, we 
present a fast method for 
inner product calculation of two tensors in the same size represented in tensor-ring format, which will be crucial for 
subsequent computations. When all the tensor ring cores of $\mathcal{T}_1$ and $\mathcal{T}_2$ are available, like 
$\{{\mathcal{G}}_d^{(1)}\in\mathbb{R}^{R_1\times I \times R_1}\}_{d=1}^D$ and $\{{\mathcal{G}}_d^{(2)}\in\mathbb{R}^{R_2\times I \times R_2}\}_{d=1}^D$, the inner product of $\mathcal{T}_1$ and $\mathcal{T}_2$ can be efficiently computed by contracting their TR core tensors as in Algorithm \ref{alg:tensor_ip}. The overall computation requires $\mathcal{O}(I R_1^2 R_2^2+(D-1)I R_1^4 R_2^4+R_1R_2)$ operations. 

\begin{algorithm}[t!]
   \caption{Inner product of two tensors represented in tensor-ring format.}
    \label{alg:tensor_ip}
    \textbf{Input:} TR cores of tensor $\mathcal{T}_1$ and $\mathcal{T}_2$ represented in TR format: $\{{\mathcal{G}}_d^{(1)}\}_{d=1}^D$ and $\{{\mathcal{G}}_d^{(2)}\}_{d=1}^D$. 
       \begin{algorithmic}[1]
       \STATE $\begin{aligned} & res \leftarrow \operatorname{einsum}\left({ }^{\prime} i n j, x n y \rightarrow i x j y^{\prime}, {\mathcal{G}}_1^{(1)}, {\mathcal{G}}_1^{(2)}\right) \end{aligned}$
       \FOR{$d=2$ to $D$}
        \STATE $\begin{aligned} & res \leftarrow \operatorname{einsum}({ }^{\prime} i x j y, j k l,y k m \rightarrow i x l m^{\prime} ,\end{aligned}$\\$\begin{aligned} & res , {\mathcal{G}}_d^{(1)}, {\mathcal{G}}_d^{(2)}  ) \end{aligned}$
       \ENDFOR
       \STATE $res$ $\leftarrow$ $\operatorname{reshape}$($res$,$[ix,lm]$) \\
       \STATE $res$ $\leftarrow$ $\operatorname{Trace}$($res$)
\end{algorithmic}
\textbf{Output:} $res$.
\end{algorithm}

\section{Mixture Tensor Ring Density Method}
    \label{sec:method}
    In this section, we begin by developing a fundamental version of the tensor ring density estimator. This version enables exact sampling, calculation of the cumulative density function, conditional probability density function, and precise computation of the partition function.
     Next, to delve into the underlying structure of the data more deeply, we introduce a mixture model of the basic version. This involves adaptive learning of the weights of different circular permutation candidates. Finally, 
     the learning methods employed are presented. 
	\subsection{Basic Model of Tensor Ring Density Estimator}
For density estimation, 
 suppose we are given independent identically distributed samples $\boldsymbol{x}^{(1)},\ldots,\boldsymbol{x}^{(N)} $, where $\boldsymbol{x}^{(n)} \in \mathbb{R}^D, n = 1, \ldots, N$ are from an unknown probability density function $p(\boldsymbol{x})$. The primary objective is to find a suitable approximation to $p(\boldsymbol{x})$. This can be achieved by obtaining an approximation $q_{\boldsymbol{\theta}}(\boldsymbol{x})$ that is close to $p(\boldsymbol{x})$ as 

\begin{equation}
 {p}(\boldsymbol{x})\approx q_{\boldsymbol{\theta}}(\boldsymbol{x}).
\end{equation}

To achieve a suitable approximation, 
structured assumptions are 
commonly employed, and the model is then optimized with respect to the parameters $\boldsymbol{\theta}$, based on these structured assumptions.  

Before we delve into the multivariate setting, it is instructive to examine the univariate case :
\begin{equation}
q_{\theta}(x) = \left\langle\boldsymbol{\alpha}, {\Phi}(x)\right\rangle = \sum_{k=1}^K \boldsymbol{\alpha}^k f_k(x),
\end{equation}
where $K$ represents the number of intervals, and $\boldsymbol{\alpha}$ is a coefficient vector comprising $K$ elements, $f_k(x)$ is the basis function of interval $k$ for interpolation. Each element in the vector $\boldsymbol{\alpha}$ is commonly referred to as a `control point’ and plays a crucial role in determining the amplitudes of the basis functions $f_k(x)$ within their respective intervals. 

Various basis functions can be used for this interpolation, including piecewise polynomials, orthogonal functions, radial basis functions, and more. In our study, we opt for uniform B-splines with equidistant knots. This choice offers several advantages, including smoothness, flexibility, non-negativity throughout the domain, and ease of initialization and computation.

In the multivariate case, we are interested in obtaining an approximation for the underlying probability distribution as follows. For simplicity, the number of intervals for every dimension is equal to $K$:
\begin{equation}
\begin{aligned}
& q_{\boldsymbol{\theta}}{(\boldsymbol{x})}=\sum_{k_1=1}^{K} \cdots  \sum_{k_D=1}^{K} \mathcal{A}^{k_1, \ldots, k_D} \mathcal{F}^{k_1, \ldots, k_D}(\boldsymbol{x}) \\
& =\sum_{k_1=1}^{K} \cdots  \sum_{k_D=1}^{K} \mathcal{A}^{k_1, \ldots, k_D} \boldsymbol{f}\left(x_1\right) \circ \cdots \circ  \boldsymbol{f}\left(x_D\right),
\end{aligned}
\end{equation}
where $\mathcal{A}\in\mathbb{R}^{K\times\ldots\times K}$ is the coefficient
tensor of order-$D$.
Here we decompose the tensor $\mathcal{A}$ by TR format:
\begin{equation}
\begin{aligned}
& \mathcal{A}\left(k_1, k_2, \ldots, k_D\right) \\
&= \operatorname{Trace}\left(\mathcal{G}^{(\mathcal{A})}_1\left(:, k_1, :\right) \mathcal{G}^{(\mathcal{A})}_2\left(:, k_2, :\right) \ldots \mathcal{G}^{(\mathcal{A})}_D\left(:, k_D, :\right)\right).
\end{aligned}
\end{equation}

$\boldsymbol{\Phi}(\boldsymbol{x})$ is a rank-1 feature map computed by
\begin{equation}
\Phi(\boldsymbol{x})=\boldsymbol{f}\left(x_1\right) \circ \cdots \circ \boldsymbol{f}\left(x_D\right),
\end{equation}
with
\begin{equation}
\boldsymbol{f}(x_d)=({f_1(x_d), \ldots, f_K(x_d)}).
\end{equation}

Each univariate is a linear combination of $K$ basis functions $\{f_{k_d}(x_d)\}_{k_d=1}^K$.
For each dimension $d = 1, \ldots D $, univariate function can be written as:
\begin{equation}
 \mathbf{Q}_d\left(x_d\right)=\sum_{k_d=1}^{K} f_{k_d}\left(x_d\right) \mathcal{G}_d^{\left(\mathcal{A}\right)}(:, k_d, :).
 \label{matrix f}
\end{equation}
For the uniform third-order quadratic B-splines used in this work, due to the local support characteristics of B-splines \cite{gordon1974b}, only 3 basis functions affect the function values of each interval. 
So Eq. \eqref{matrix f} can be rewritten as:
\begin{equation}
\begin{aligned}
&\mathbf{Q}_d(x_d) = f_{k_d-1}(x_d) \mathcal{G}_d^{(\mathcal{A})}(:, k_d-1, :) \\
&+ f_{k_d}(x_d) \mathcal{G}_d^{(\mathcal{A})}(:, k_d, :) + f_{k_d+1}(x_d)
\mathcal{G}_d^{(\mathcal{A})}(:, k_d+1, :).  
\end{aligned}
\end{equation}
In this case, $\mathcal{G}_d^{\left(\mathcal{A}\right)}\in\mathbb{R}^{R_{d-1} \times K \times R_d}$ is a 3-dimensional TR core of coefficient tensor in TR format. For each dimension, every  univariate function  $\mathbf{Q}_d\left(x_d\right) \in\mathbb{R}^{R_{d-1}  \times R_d}$ is a matrix-valued function. 

     \begin{figure*}[!htbp]
        \centering
		\subfigure{\includegraphics[scale=0.4]{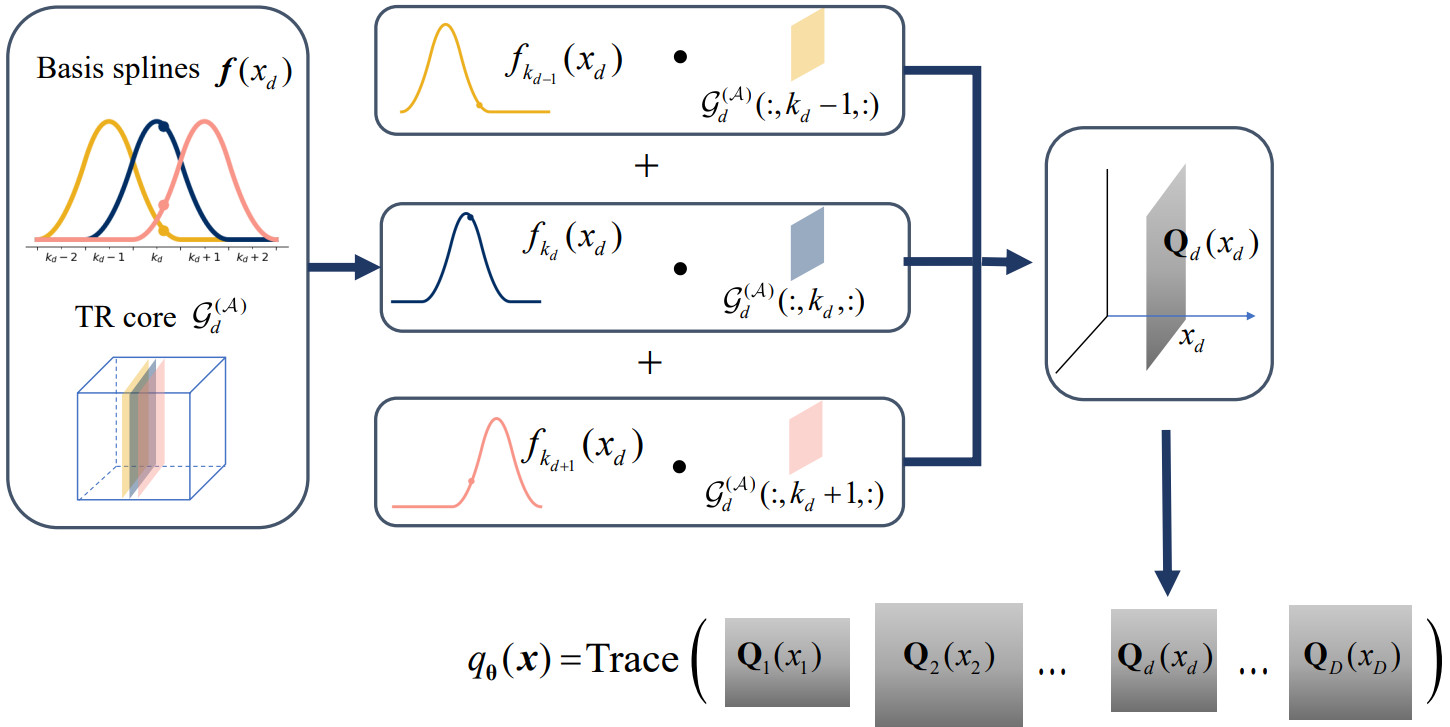}}
		\caption{The complete calculation process of $\mathbf{Q}_d\left(x_d\right)$ and calculation of $q_{\boldsymbol{\theta}}(\boldsymbol{x})$ using $D$ matrices $\{{\mathbf{Q}_d}(x_d)\}_{d=1}^D$ by Eq. \eqref{ccc}. For $d\in D,  \mathbf{Q}_d\left(x_d\right)$ is calculated by three lateral slices of TR core $\mathcal{G}^{\left(\mathcal{A}\right)}_d$: $\mathcal{G}_d^{\left(\mathcal{A}\right)}(:, k_d-1, :)$, $\mathcal{G}_d^{\left(\mathcal{A}\right)}(:, k_d, :)$, $\mathcal{G}_d^{\left(\mathcal{A}\right)}(:, k_d+1, :)$ and its corresponding  B-spline basis function that takes a non-zero value over $k_d$ interval.    } 
        \label{full}
	\end{figure*}
 
Under normal circumstances, a lower rank implies stronger independence between variables. When the TR rank of this TR model equals 
to $1$, the variables become completely independent. In this case, the TR format then degenerates to rank one CP decomposition, which is the outer product of vectors, each of length $K$.

For a given $D$ dimensional sample $\boldsymbol{x}\in\mathbb{R}^{D}$, we can calculate its probability by calculating trace after a series of matrix products:
\begin{equation}
q_{\boldsymbol{\theta}}(\boldsymbol{x})=\operatorname{Trace}{(\mathbf{Q}_1\left(x_1\right) \mathbf{Q}_2\left(x_2\right) \cdots \mathbf{Q}_D\left(x_D\right))}.
\label{ccc}
\end{equation}

Fig.\ref{full} shows the complete calculation diagram of $\mathbf{Q}_d\left(x_d\right) \in\mathbb{R}^{R_{d-1}  \times R_d}$ and calculation of $q_{\boldsymbol{\theta}}(\boldsymbol{x})$ using $D$ matrices $\{{\mathbf{Q}_d}(x_d)\}_{d=1}^D$ by Eq. \eqref{ccc}.

    \subsection{Marginal probability density and its derivatives}
In this section, we will explain how to compute the marginal probability density and its derivatives within our model. One of the key advantages of the tensor-based density estimation method is its ability to accurately calculate the cumulative probability density, marginal probability density, and conditional probability density. The structure of the tensor ring format allows us to conveniently marginalize specific dimensions by summing all lateral slices corresponding to dimension $d \in P$, where $P$ represents the set of indices to be marginalized:
\begin{equation}
\mathcal{G}^{(\mathcal{C})}_d=\sum_{k_d=1}^{K} \mathcal{G}^{\left(\mathcal{A}\right)}_d(:,k_d,:).   
\end{equation}
The complete calculation process is outlined in Algorithm \ref{alg:B}.
    \begin{algorithm}[!t]
   \caption{Calculate marginal density coefficient tensor $\mathcal{C}$ in TRDE.}
    \label{alg:B}
    \textbf{Input:} All TR cores $\{{\mathcal{G}_d}^{(\mathcal{A})}\}_{d=1}^D$ of the coefficient tensor ${\mathcal{A}}$, a set $P$ of indices that are marginalized.
       \begin{algorithmic}[1]
       \FOR {$d=1$ to $D$}         \IF{$d \in P$}
          \STATE $\mathcal{G}^{(\mathcal{C})}_d=\sum_{k_d=1}^{K} \mathcal{G}^{\left(\mathcal{A}\right)}_d(;,k_d,;);$
         \ELSE
          \STATE $\mathcal{G}^{(\mathcal{C})}_d=\mathcal{G}^{\left(\mathcal{A}\right)}_d$
         \ENDIF
       \ENDFOR
    \STATE $\mathcal{C}\left(k_1,\ldots,k_D\right)=\operatorname{Trace}\left(\prod_{d=1}^D \mathcal{G}^{\left(\mathcal{C}\right)}_d\left(k_d\right)\right)$
\end{algorithmic}
\textbf{Output:} Cumulative probability density coefficient tensor $\mathcal{C}$ of a variable set $P$ are marginalized.
\end{algorithm}
Overall, the whole Algorithm \ref{alg:B} requires $\mathcal{O}((K^2(D-P)+K+P)R^3+(K-1)PR^2+R)$ operations. $R$ is the TR rank of TR model, $D$ is the number of dimensions, $P$ is the number of dimensions that need to be marginalized, and $K$ is the number of intervals of each dimension. 
The cumulative probability density can be easily obtained by changing the upper limit of the definite integral of B-splines:
\begin{equation}
\begin{aligned}
& q_{\boldsymbol{\theta}}\left(x_1, \ldots, x_{d-1}, x_d<a, \ldots, x_D\right) \\
& =\langle\mathcal{A}, \Phi\left(x_1, \ldots, x_{d-1}\right) \circ \int_{-\infty}^a f\left(x_d\right) d x_d \\
& \circ \Phi\left(x_{d+1}, \ldots, x_D\right)\rangle,
\end{aligned}
\end{equation}
while the conditional probability density function can be calculated by classical Bayesian theory.

\subsection{Sampling}

Sampling is a crucial application of probability density estimation. Generative Adversarial Networks (GANs) and Variational Autoencoders (VAEs) are renowned for their exceptional sampling capabilities. These models can generate highly realistic image samples using forward-pass networks without the need for explicit real density function calculations. In contrast, energy density models require additional techniques to obtain approximate samples, such as Markov Chain Monte Carlo sampling or rejection sampling.

Recently, dimension-wise sampling based on TT format (\cite{dolgov2020approximation,cui2022deep}) has proven remarkably successful, primarily due to its dimensionally linear storage and computational costs. Inspired by this, we have developed an autoregressive sampling method based on the TR format. This sampling algorithm is applied to our probability density estimator.

The autoregressive sampling technique simplifies the process of sampling a $D$-dimensional random vector by breaking it down into a sequence of $D$ univariate, one-dimensional sampling tasks. Leveraging the inherent smoothness of B-splines in our model, we adopt a straightforward and uncomplicated sampling approach by employing the inverse transformation in an autoregressive fashion, as outlined below.

First, produce a random seed $\boldsymbol{u}\sim\mathcal{U}([0 ; 1]^D)$. The coordinates of each dimension are then sequentially sampled: $q_{\boldsymbol{\theta}}\left(\xi_1<x_1\right)=u_1$, $q_{\boldsymbol{\theta}}(\xi_2<x_2 \mid\xi_1=x_1)=u_2, \ldots$.
    Sampling in each dimension requires seven substeps:
\begin{enumerate}[1.]
\item Precompute the TR cores $\{{\mathcal{G}}^{\left(\mathcal{C}\right)}\}_{d=1}^D$ 
of the cumulative density  $\int q_{\boldsymbol{\theta}}\left(\boldsymbol{x}\right) d \boldsymbol{x}$  using TR cores of coefficient tensor $\mathcal{A}$ by Algorithm \ref{alg:B}.
\end{enumerate}  
\begin{enumerate}[2.]
\item Contract all inner indices of ${q}_{\boldsymbol{\theta}}\left(x_1, \ldots, x_{d-1}\right)$ with cores ${\mathcal{G}}^{\left(\mathcal{A}\right)}_1,\ldots, {\mathcal{G}}^{\left(\mathcal{A}\right)}_{d-1}$, we call it $\mathbf{Q}_d^{\text {left }}$, which is a matrix of size $[R_{D}, R_{d-1}]$.
\end{enumerate}  
\begin{enumerate}[3.]
\item Contract all inner indices of the integral $\int q_{\boldsymbol{\theta}}\left(x_{d+1}, \ldots, x_D\right) d x_{d+1} \ldots d x_D$ with cores ${\mathcal{G}}^{(\mathcal{C})}_{d+1},$$\ldots{\mathcal{G}}^{(\mathcal{C})}_{D}$, resulting in a matrix denoted as $\mathbf{Q}_d^{\text {right }}$, with dimensions $[R_{d},R_{D}]$.
\end{enumerate}  
\begin{enumerate}[4.]
\item Contraction of ${\mathcal{G}}^{(\mathcal{A})}_{d}$ with $\mathbf{Q}_d^{\text {left }}$ and $\mathbf{Q}_d^{\text {right }}$ along left and right indexes. The result is a third order tensor $\mathcal{I}_d$ of shape $[R_{D},K,R_{D}]$.
\end{enumerate}  
\begin{enumerate}[5.]
\item perform trace of every lateral slice of $\mathcal{I}_d$, which results in a one-dimensional basis vector $\boldsymbol{q}_d^{\text {inner }}$ of size $K$, where $K$ is the number of intervals.
\end{enumerate}  
\begin{enumerate}[6.]
\item Find $x_d$ satisfy $\langle \boldsymbol{q}_d^{\text {inner }},\int_{-\infty}^{x_d} f\left(\xi_d\right) d \xi_d \rangle =u_d$. Calculate $\mathbf{Q}_d(x_d)$ using B-spline interpolation as shown in  Fig.\ref{full}. It is a matrix sized of $[R_{d-1},R_{d}]$. 
\end{enumerate}  
\begin{enumerate}[7.]
\item Update $\mathbf{Q}_{d+1}^{\text {left }}=\mathbf{Q}_d^{\text {left }}\mathbf{Q}_d(x_d)$.
\end{enumerate}  

Note that the sixth step involves a one-dimensional search on the cumulative probability function, which is a monotonically increasing function. Various searching algorithms can be employed, such as binary search, golden section search, or Fibonacci search. The complete process is summarized in Algorithm \ref{sample}. 

Next, we perform computational complexity analysis of Algorithm \ref{sample}. We emphasize that this algorithm works for unequal TR ranks. In 
Algorithm \ref{sample}, Step 4 requires $\mathcal{O}(2KR^3)$ operations for matrix multiplication. Step 5 needs $\mathcal{O}(KR)$ operations for $\text{trace}$ calculation. Step 6 needs $\mathcal{O}(K)$ operations for vector inner product and $\mathcal{O}(3R^2)$ operations for matrix summations in interpolation. Step 7 needs $\mathcal{O}(R^3)$ operations for matrix multiplication. In general, Algorithm \ref{sample}  needs $\mathcal{O}(D((2K+1)R^3+3R^2+KR+K))$ operations. 
It is worth noting that the calculation of $\{{\mathcal{G}}^{\left(\mathcal{C}\right)}\}_{d=1}^D$, $\mathbf{Q}_d^{\text {left }}$ and $\mathbf{Q}_d^{\text {right }}$  can be calculated on-the-fly. Therefore, their computational complexity is not a limiting factor for Algorithm \ref{sample}. In addition, the computational complexity of a one-dimensional search depends on the selected search algorithm and its accuracy.

    \begin{algorithm}[!t]
   \caption{Obtain an exact sample from tensor ring estimator 
   $\boldsymbol{q}_{\boldsymbol{\theta}}\left(\boldsymbol{x}\right)$.}
    \label{sample}
    \textbf{Input:}  Sample seed $\boldsymbol{u}$ from $\boldsymbol{u} \sim U\left([0 ; 1]^D\right)$,
 TR cores $\{{\mathcal{G}}^{\left(\mathcal{C}\right)}\}_{d=1}^D$ of the marginal density $\int{q}_{\boldsymbol{\theta}}\left(\boldsymbol{x}\right)d{\boldsymbol{x}}$, and TR cores $\{{\mathcal{G}}^{\left(\mathcal{A}\right)}\}_{d=1}^D$ of the coefficient tensor ${\mathcal{A}}_{\boldsymbol{\theta}}$.
       \begin{algorithmic}[1]
       \STATE Initialize $\mathbf{Q}_d^{\text {right }}$ with identity matrix of shape $[R_{D},R_{D}]$
       \FOR{$d=D$ to $2$}
         \STATE  Precompute $\mathbf{Q}_{d-1}^{\text {right }}=\sum\limits_{k_{d}}\sum\limits_{r_{d}} \mathcal{G}_{d-1}^{\left(\mathcal{C}\right)} \cdot \mathbf{Q}_{d}^{\text {right }}$
       \ENDFOR
       \STATE Initialize $\mathbf{Q}_d^{\text {left}}$ with identity matrix of shape $[R_{D},R_{D}]$
       \FOR{$d=1$ to $D$}
         \STATE Compute $\mathcal{I}_{d}=\sum\limits_{r_{d-1}}\sum\limits_{r_{d}} \mathbf{Q}_{d}^{\text {left }}{\mathcal{G}}_{d}^{\left(\mathcal{A}\right)} \mathbf{Q}_{d}^{\text {right }}$
         \STATE Compute $\boldsymbol{q}_d^{\text {inner }}=\operatorname{Trace}$ lateral slices of three-order tensor  $\mathcal{I}_{d} $
         \STATE Find $x_d$ such that $\langle \boldsymbol{q}_d^{\text {inner }}, \int_{-inf}^{x_d} {{f}}\left(\xi_d\right) d \xi_d\rangle=u_d$,
         \STATE Update $\mathbf{Q}_{d+1}^{\text {left }}=\mathbf{Q}_d^{\text {left }}{\mathbf{Q}}_{d}(x_d)$
       \ENDFOR
\end{algorithmic}
\textbf{Output:} $q_{\boldsymbol{\theta}}$-distrbuted sample vector $\boldsymbol{x}$.
\end{algorithm}
	\subsection{Mixture model for  basic estimator}	 
In recent years, significant research efforts have been dedicated to Tensor Network Structure Search (TN-SS) \cite{li2022permutation} and Tensor Network Permutation Search (TN-PS) \cite{li2020high}. These studies have provided empirical evidence highlighting the substantial influence of feature permutations on the potential structure of learned tensors. We firmly believe that exploring these suboptimal tensor network structures offers a promising avenue to enhance our understanding of the underlying data.

In contrast to the common strategy of exclusively seeking optimal permutations based on data characteristics, our approach draws inspiration from ensemble learning and the classical Gaussian mixture model. Through this approach, we extend the foundational TRDE into a mixture format, unlocking new possibilities for modeling the underlying data distribution.

Our approach embraces diverse suboptimal tensor structures, acknowledging their potential to capture nuanced patterns and hidden information in the data. This diversification empowers us to explore a spectrum of potential structures, moving beyond the constraints of a single `rigid' tensor network. The resulting mixture format enhances the flexibility and adaptability of our model, making it more robust in capturing complex, multifaceted data distributions. As a result, we extend the basic tensor ring estimator to a mixture format, as described below:
\begin{equation}
q_{\boldsymbol{\theta}}(\boldsymbol{x})=\sum_{m=1}^{M}{\sigma_m}\sum_{k_1=1}^{K} \cdots  \sum_{k_D=1}^{K} \mathcal{A}_m^{k_1, \ldots, k_D} \mathcal{F}_{k_1, \ldots, k_D}(\boldsymbol{x}).
\label{mix}
\end{equation}
The model can be conceptually interpreted as a Bayesian model, in which ${\sigma}$ can be seen as a `hidden variable`. In this context, $\boldsymbol{\sigma}$ represents the weight vector, where the sum of its elements equals 1. Each component of the weight vector corresponds to a different permutation of the TR.

\begin{figure}[!t]
        \centering
		\subfigure{\includegraphics[scale=0.7]{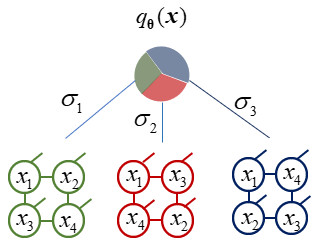}}
		\caption{A schematic diagram of a four-dimensional TERM, composed of $\frac{(4-1)!}{2}$ basic TRDE components with adaptive weight, collectively forming TERM.} 
        \label{TERMM}
	\end{figure}

For clarity, Fig.\ref{TERMM} illustrates a four-dimensional TERM diagram. In this diagram, the weight vector $\boldsymbol{\sigma}$ embodies the idea that different permutations of the tensor ring play a role in modeling the data distribution. By allowing for these permutations, the model captures a range of potential data structures, enhancing its adaptability and capability to represent complex data distributions.

Due to the TR's circular dimensional permutation in-variance, the number of different permutation candidates is $1/D$ times that of the tensor train format, without incurring any additional storage costs\cite{li2022permutation}. $D$ is the dimension of the data.
\subsection{Learning and training}
Without any prior knowledge, we do not know the true probability function $\boldsymbol{p(\boldsymbol{x})}$. Benefit from the tractable partition function, we could train the model by maximizing the log-likelihood function:
\newline
\begin{equation}
\label{logcal}
\begin{aligned}
 &\sum_{i=1}^{N} \log \frac{\boldsymbol{q}_{\boldsymbol{\theta}}(\boldsymbol{x}_i)}{\int \boldsymbol{q}_{\boldsymbol{\theta}}(\boldsymbol{x})d\boldsymbol{x}}\\
 &= \sum_{i=1}^{N} \left(\log {\boldsymbol{q}_{\boldsymbol{\theta}}(\boldsymbol{x}_i)}- \log\left({\int \boldsymbol{q}_{\boldsymbol{\theta}}(\boldsymbol{x})d\boldsymbol{x}}\right)\right).
\end{aligned}
\end{equation}
where the non-normalized probability ${\boldsymbol{q}_{\boldsymbol{\theta}}(\boldsymbol{x}_i)}$ is referred to as the `positive phase', while the normalized factor ${\int \boldsymbol{q}_{\boldsymbol{\theta}}(\boldsymbol{x})d\boldsymbol{x}}$ is known as the `negative phase' or partition function.

Non-negativity of the probability density function is essential for our model. This means that for any $\boldsymbol{x}$, the following condition must hold
\begin{equation}
\forall \boldsymbol{x}, \langle\mathcal{A}, \Phi(\boldsymbol{x})\rangle \geq 0.
\end{equation}
where B-splines are naturally non-negative. However, we still need to ensure that the coefficients tensor $\mathcal{A}$ is strictly positive. 
To tackle this issue, we utilize the squared variant similar to a series of work based on TT \cite{dolgov2020approximation,cui2022deep,novikov2021tensor} format:
\begin{equation}
\boldsymbol{q}_ {\boldsymbol{\theta}} (\boldsymbol{x})=\left\langle\mathcal{A}, \Phi(\boldsymbol{x})\right\rangle^2.
\end{equation}
For our basic estimator, log-likelihood function's positive phase is convenient to compute. The negative phase $\int\left\langle\mathcal{A}, \Phi(\boldsymbol{x})\right\rangle^2 d \boldsymbol{x}$ can be calculated by as follows:
\begin{equation}
\begin{aligned}
\int\left\langle\mathcal{A}, \Phi(\boldsymbol{x})\right\rangle^2 d \boldsymbol{x} & =\int\left\langle\mathcal{A},(\Phi(\boldsymbol{x})  \circ\Phi(\boldsymbol{x})) \mathcal{A}\right\rangle d \boldsymbol{x} \\
& =\left\langle\mathcal{A},\left(\int \Phi(\boldsymbol{x}) \circ \Phi(\boldsymbol{x}) d \boldsymbol{x}\right) \mathcal{A}\right\rangle.
\end{aligned}
\end{equation}
\begin{equation}
\begin{aligned}
& {\left[\int \Phi(\boldsymbol{x})\circ\Phi(\boldsymbol{x}) d \boldsymbol{x}\right]} \\
& =\int \Phi(x_1, \ldots, x_D) \circ\Phi(x_1, \ldots, x_D)d x_1 \cdots d x_D \\
& =\int (\boldsymbol{f}_{1}\left(x_1\right) \circ \boldsymbol{f}_{1}\left(x_1\right))\circ(\boldsymbol{f}_{2}\left(x_2\right) \circ \boldsymbol{f}_{2}\left(x_2\right)) \\
&\ldots (\boldsymbol{f}_{D}\left(x_D\right) \circ \boldsymbol{f}_{D}\left(x_D\right)) d x_1 \cdots d x_D \\
& =\int (\boldsymbol{f}_{1}\left(x_1\right) \circ \boldsymbol{f}_{1}\left(x_1\right))d x_1
 \circ\ldots  \\
&\circ\int(\boldsymbol{f}_{D}\left(x_D\right) \circ \boldsymbol{f}_{D}\left(x_D\right)) d x_D\\
& =\mathbf{M}_{1}  \circ\ldots \circ \mathbf{M}_{D}.
\end{aligned}
\end{equation}
For a uniform third-order quadratic B-spline, $\left\{{\mathbf{M}_{d}}\right\}_{d=1}^D$ are five-diagonal matrices, we call it mass matrices which are all symmetric positive definite. As mentioned above, tensor $\left[\int \Phi(\boldsymbol{x}) \circ \Phi(\boldsymbol{x}) d \boldsymbol{x}\right]$ is a rank-1 tensor by outer product of $D$ matrices $\left\{{\mathbf{M}_{d}}\right\}_{d=1}^D$. The calculation of negative phase of the log-likelihood function is shown in Algorithm \ref{negcal}.
In Algorithm \ref{negcal}, the computation of each dimension $d\in D$ requires $\mathcal{O}({K^2}{R^2})$ operations for the `einsum' calculation. For whole Algorithm \ref{negcal}, it needs $\mathcal{O}((D-1)KR^8+KR^4+(K^2+1)R^2)$ operations.
    \begin{algorithm}[!t]  
		\caption{Calculation the negative phase of (\ref{logcal})}  
		\label{negcal}
		\textbf{Input:}  matrices $\{\mathbf{M}_{d}\in \mathbb{R}^{1 \times K \times K \times 1}\}_{d=1}^D$,  TR cores $\{{\mathcal{G}}^{\left(\mathcal{A}\right)}\}_{d=1}^D$ of the coefficient tensor ${\mathcal{A}}$
		\begin{algorithmic}[1]   
			\FOR {$d=1$ to $D$}
            \STATE $ \mathcal{B}_d \leftarrow \operatorname{einsum}({ }^{\prime} rms, tmnu \rightarrow rtnsu^{\prime}, \mathbf{M}_{d}, \mathcal{G}_d^{\left(\mathcal{A}\right)})$
            \ENDFOR
            \STATE $\mathcal{B}_d$ $\leftarrow$
            $\operatorname{reshape}$($\mathcal{B}_d$,$[rt,n,su]$)
            \FOR {$d=1$ to $D$}
            \STATE Inner product of TR cores of $\mathcal{A}_{\boldsymbol{\theta}}$: $\{{\mathcal{G}}^{\left(\mathcal{A}\right)}\}_{d=1}^D$ and $D$ TR cores : $\{{\mathcal{B}_d}\}_{d=1}^D$ using $\text { Algorithm } 1$
            \ENDFOR                           
		\end{algorithmic} 
		\textbf{Output:} $res$. 
	\end{algorithm}

  Similarly, for TERM, we can maximize the log-likelihood as
\begin{equation}
\sum_{i=1}^N \log (\frac{\sum_{m=1}^M{\boldsymbol{q}_{\boldsymbol{\theta}_m}\left(\boldsymbol{x}_i\right)}}{{\sum_{m=1}^M{\int \boldsymbol{q}_{\boldsymbol{\theta}_m}(\boldsymbol{x}) d \boldsymbol{x}}}} ).
\end{equation}
In the log-likelihood function, the weights of various permutation components, denoted as $\sigma_m$, are dynamically adjusted through adaptive updates, considering the magnitudes of their respective partition functions. Following the training process, the trained values of $\sigma_m$ can be computed  by:
\begin{equation}
\sigma_m=\frac{\int \boldsymbol{q}_{\boldsymbol{\theta}_m}(\boldsymbol{x}) d \boldsymbol{x}}{{\sum_{m=1}^M{\int \boldsymbol{q}_{\boldsymbol{\theta}_m}(\boldsymbol{x}) d \boldsymbol{x}}}} .
\end{equation}

\section{Experiment}
In this section, we begin by evaluating the effectiveness of our density estimation and sampling method on 2-D toy distributions. Our objective is to visually illustrate the feasibility of the method and the influence of each parameter on probability density estimation. 
Following this, we further conduct sampling experiments on a 3D synthetic dataset to verify the superiority of our algorithm in both estimation and sampling and its generalization to high-order distributions. At last, real datasets are employed for comparison experiments with state-of-the-art methods.
  All experiments were conducted using a single NVIDIA V100 GPU in conjunction with an Intel Xeon E5-2698 V4 CPU operating at 2.2GHz.
 \subsection{Toy 2-D Distributions}
In the first stage of our evaluation, we consider several commonly used two-dimensional toy datasets. Actually, for 2-D distributions, the low-rank TR decomposition degenerates to low-rank matrix approximation and there is no permutation selection problem. However, due to the visibility and ease of analysis of the 2-D distributions, here we exploit them to verify the feasibility and parameter sensitivity of the proposed TRDE.

 \subsubsection{Density Estimation}

\begin{figure}[!t]
     \begin{center}
    \includegraphics[scale=0.3]{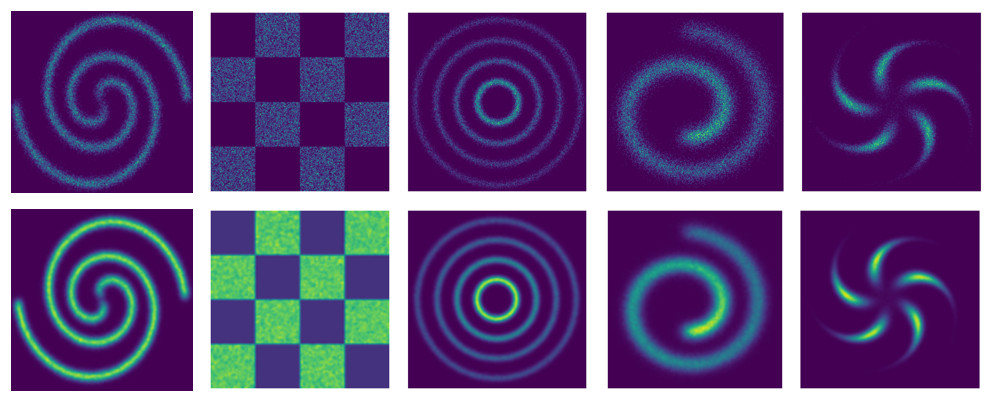}
    \caption{Approximation of different 2-D toy distributions. The top row is real data, and the bottom row is the result of TRDE. From left to right: 2-Spirals Checkerboard, Rings, Swissroll, Pinwheel.} 
    \label{toy}
    \end{center}
\end{figure}

First, we employ the proposed basic estimator TRDE to approximate the desired 2-D distributions. The TR rank is set to 12, and the basis size is set to 256. As depicted in Fig.~\ref{toy}, our method proves to be highly effective,  nearly perfectly matching the distributions of different complex 2-D datasets. There are no multimodal or discontinuous problems that are often encountered in probability density estimators based on neural networks.

\subsubsection{The Effect of Parameters}
\begin{figure}[!t]
\centering
\includegraphics[scale=0.35]{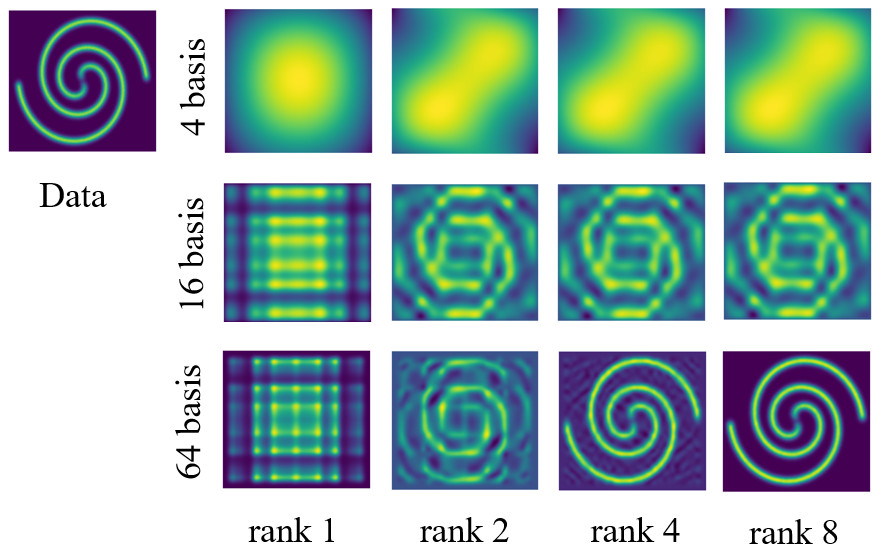}
    \caption{Approximation of the `2 Spirals' distribution using TRDE with various basis function sizes and TR-ranks.} 
    \label{hyper}
\end{figure}

In our model, each parameter is highly interpretable: the basis size is analogous to the `bin' size of a histogram, while the TR rank and the number of components in the mixture model indicate the expressive power of the model.
We conduct experiments on the 2-spirals dataset, varying hyperparameter settings. Specifically, we vary the basis size in $\{ 4, 16, 64\}$, and the TR rank in $\{1, 2, 4, 8\}$. Fig.~\ref{hyper} illustrates the impact of these hyperparameters in our proposed model. In the context of a 2-D distribution, the TR rank is equivalent to the rank of a matrix. It can be observed that higher TR ranks in the approximation result in reduced noise. An increase in the number of basis functions leads to improved resolution. 

 \subsubsection{Sampling performance comparison}
Here we compare the sampling performance of our proposed TRDE with Real NVP\cite{dinh2016density}, Neural Spline Flow (NSF)\cite{dolgov2020approximation}, the state-of-the-art normalizing flow model in a medium-dimensional setting. Four two-dimensional datasets are included: Two spirals, Checkerboard, Tree, and Sierpinski Carpet. To be fair, we set the parameters of the three models to the same order of magnitude. We use 4 layers of affine layer's Real NVP, 2  layers of rational quadratic coupling layer's NSF, and TRDE's TR rank is set to 8, and the basis size set to 128. 
On account of approximation to the original distribution, our method has fewer parameters than neural networks.

\begin{figure}[!t]
\centering
\includegraphics[scale=0.5]{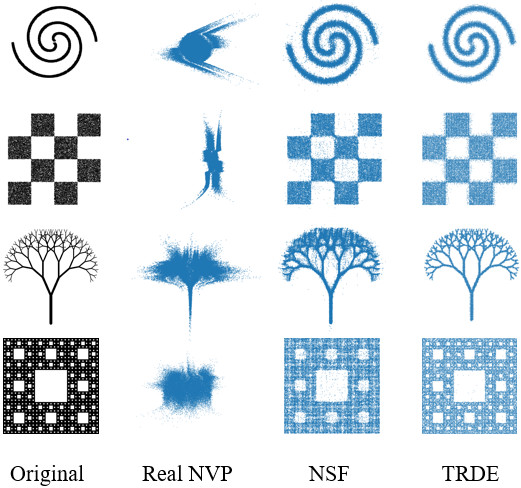}
    \caption{ Sampling performance comparison of TRDE and Real-NVP, NSF on 2-dimensional toy distributions. 
} 
    \label{diff}
\end{figure}

As depicted in Fig.~\ref{diff}, the results of sampling on four datasets highlight the limitations of Real NVP due to its inflexible flow structure. Notably, our model exhibits notable improvements in several aspects. In the case of the 2-Spirals and Checkerboard datasets, our model displays a reduced number of outliers. In the Tree dataset experiment, our model excels in producing more detailed samples, particularly at the extremities of the branches. Furthermore, when considering the Sierpinski Carpet dataset, it becomes evident that our model generates samples with intricate details and reduces noise at the edges of the images.

\subsection{3-D Distributions}
\begin{figure}[!t]
\centering
\includegraphics[scale=0.5]{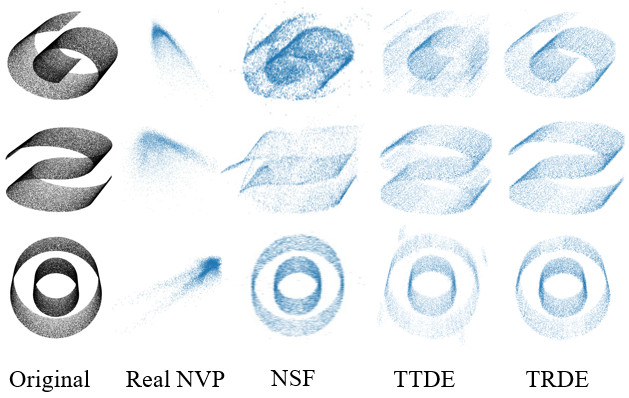}
    \caption{ Sampling performance comparison of TRDE and Real NVP, NSF, TTDE on 3-dimensional toy distributions. 
} 
    \label{3d}
\end{figure}

Next, we conduct experiments on three 3D datasets, namely Swissroll, Circles, and S-curve, to compare the sampling capabilities of Real NVP\cite{dinh2016density}, NSF\cite{durkan2019neural}, TTDE\cite{novikov2021tensor} and the proposed TRDE. Our training dataset consisted of 50,000 points, and we simultaneously sample 20,000 samples from the trained models. To ensure a fair comparison, we standardize the total number of parameters across the three models. Real NVP employed 4 affine layers, NSF utilized two piecewise rational quadratic coupling layers, TTDE's TT rank is set to 13, and TRDE's TR rank is set to 6. The results are visualized in Fig.~\ref{3d}.
When applied to the Swissroll dataset, Real NVP faces limitations due to the simplicity of its affine layer transformations, resulting in an inadequate sampling of 3-D data. In contrast, NSF generates excessive noise, and TTDE seems to omit crucial information related to a specific mode. Conversely, TRDE excels by accurately identifying the two-dimensional manifold within the three-dimensional data, effectively reconstructing the distribution.

In the case of the S-Curve dataset, NSF's handling of the edges of the low-dimensional manifold appears somewhat coarse. TTDE, while attempting to approximate the `S' shape, produces a figure more resembling the number `8'. Meanwhile, TRDE continues to outperform, showcasing its superior ability to capture the dataset's nuances.

For the Circle dataset, TTDE introduces extraneous clusters of noise around the dual circles, and NSF also shows some distant noise encircling the double circle. TRDE, however, does not exhibit these issues, demonstrating its robustness. The detailed comparison of total parameters and sampling time for these models is provided in Table~\ref{3dd}. Notably, our model stands out with the lowest parameter count and the shortest sampling time, showcasing efficiency as it doesn't require rapid sampling of decoders and uses a minimal number of parameters.

Furthermore, we calculate the Kullback-Leibler (KL) divergence between the sampled data and the original sample distribution, with the results presented in Table ~\ref{kl}. Figure ~\ref{3d} illustrates that samples generated by TRDE exhibit the lowest KL divergence across all three datasets. This indicates that the distribution approximated by TRDE is closely aligned with the original distribution, affirming its effectiveness in capturing the true characteristics of these complex datasets.

\begin{table}[ht]
		\centering
        \caption{The total number of parameters and sampling time for Real NVP,NSF, TTDE and TRDE when sampling the 3-D datasets.}
        \label{3dd}
        \scalebox{1}{
		\begin{tabular}{c|cc}
			\toprule
			&Parameter (1e3)  & Sampling time (sec)  \\
			\midrule
            Real NVP &18.63  &9.42   \\
            NSF &28.02  &11.41  \\
            TTDE &12.48 &14.25  \\
            TRDE(ours) &\textbf{6.912}  &\textbf{9.35}  \\
            \bottomrule

		\end{tabular}}
  \end{table}
  \begin{table}[ht]
		\centering
        \caption{KL Divergence Comparison for Sampled Data from Real NVP, NSF, and TRDE across 3 3D Datasets}
        \label{kl}
        \scalebox{1}{
		\begin{tabular}{c|ccc}
			\toprule
			& Circles& S curve &Swissroll  \\
			\midrule
             Real NVP &0.611 &0.242&0.127 \\
             NSF &0.053 &0.045&\textbf{0.021}  \\
             TTDE &0.023 &0.041&0.031  \\
            TRDE(ours) &\textbf{0.018}  &\textbf{0.036}&\textbf{0.021}\\
            \bottomrule

		\end{tabular}}
  \end{table}
 \subsection{Real world dataset density estimation}
 \begin{table*}[h!]
		\centering
		\caption{ The average negative log-likelihood on tabular datasets for density estimation (lower is better). 
  *On the Hepmass and Miniboone datasets, a significant issue of overfitting becomes evident like TTDE. Moreover, we observe that the TERM exacerbates this overfitting phenomenon as the number of mixture components increases. So we use the TRDE results.}.
        \label{true}
        \setlength{\tabcolsep}{8mm}
		\begin{tabular}{c|cccc}
			\toprule
			       &POWER  & GAS  & HEPMASS & MINIBOONE  \\
			\midrule
            Gaussians &7.74  &3.58  &27.93 &37.24  \\
			FFJORD &-0.46  &-8.59  &14.92 &10.43  \\ 
			Real NVP &-0.17  &-8.33  &18.71 &13.84 \\ 
            GLow &-0.17  &-8.15  & 18.92 & 11.35\\
			CP-Flow  &-0.52  &-10.36  & 16.93 &10.58  \\ 
            GF & -0.57  & -10.13& 17.59  &10.32\\ 
            RQ-NSF & -0.64  & -13.09& \textbf{14.75}  &\textbf{9.67}\\
            TTDE  &-0.46 &-8.93  &21.34&28.77 \\
            TRDE(Ours)  &-1.65&-10.17  &21.32*&28.52*  \\
            TERM (Ours) &\textbf{-1.73} &\textbf{-14.91}  &21.32* &28.52*   \\
            \bottomrule
            MAF  &-0.24 &-10.08  &17.70&11.75  \\
            MADE  &3.08 &-3.56 &20.98&15.59  \\
            BNAF  &-0.61 &-12.06  &14.71&8.95  \\
            TAN  &-0.48 &-11.19  &15.12&11.01  \\
            UMNN &-0.63  &-10.89  &13.99 &9.67  \\
            MAF-DDSF  &-0.62 &-11.96  &15.09&8.86  \\
            \bottomrule
		\end{tabular}		
	\end{table*} 
 
  \begin{figure*}[ht]	
       
        \centering
		\subfigure{\includegraphics[scale=0.20]{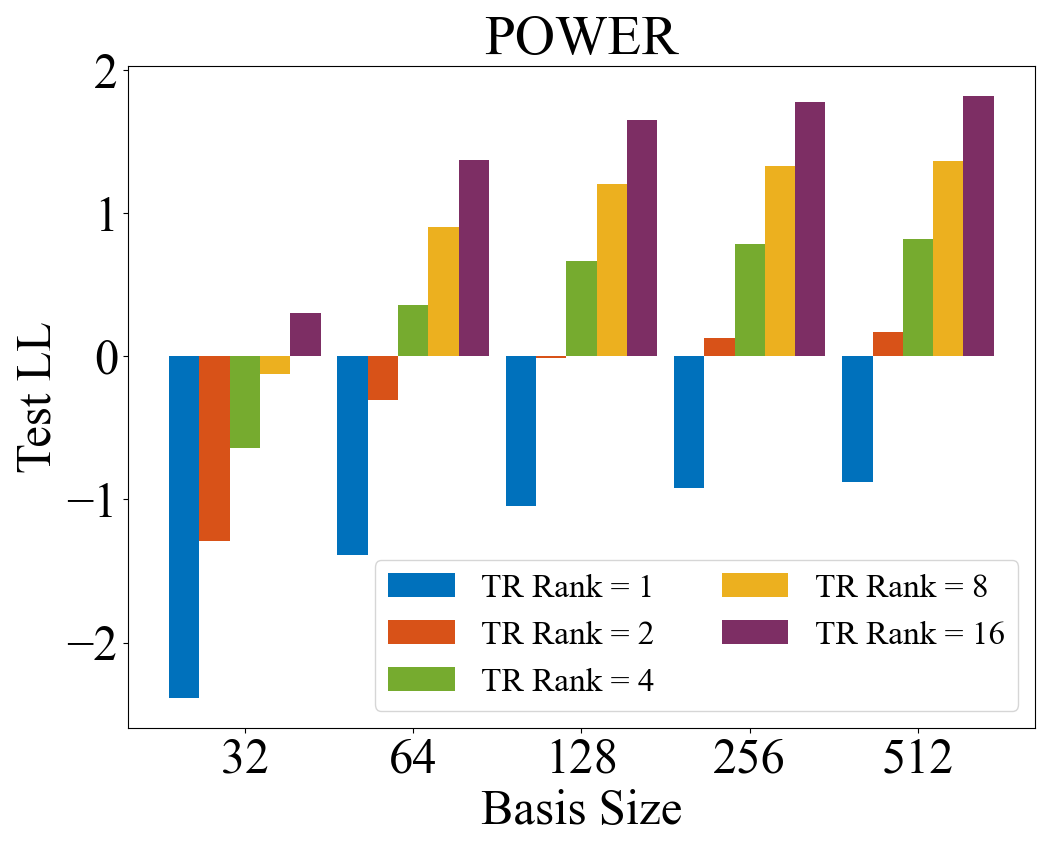}}
	    \subfigure{\includegraphics[scale=0.20]{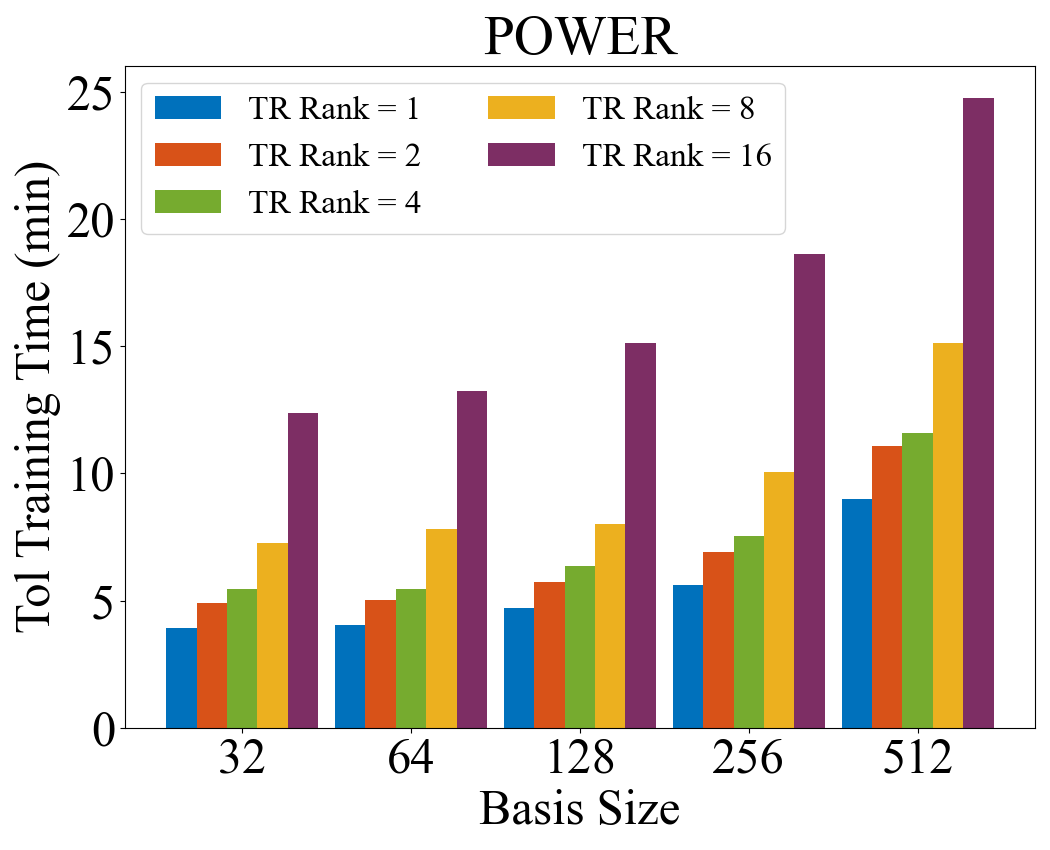}}
       \subfigure{\includegraphics[scale=0.20]{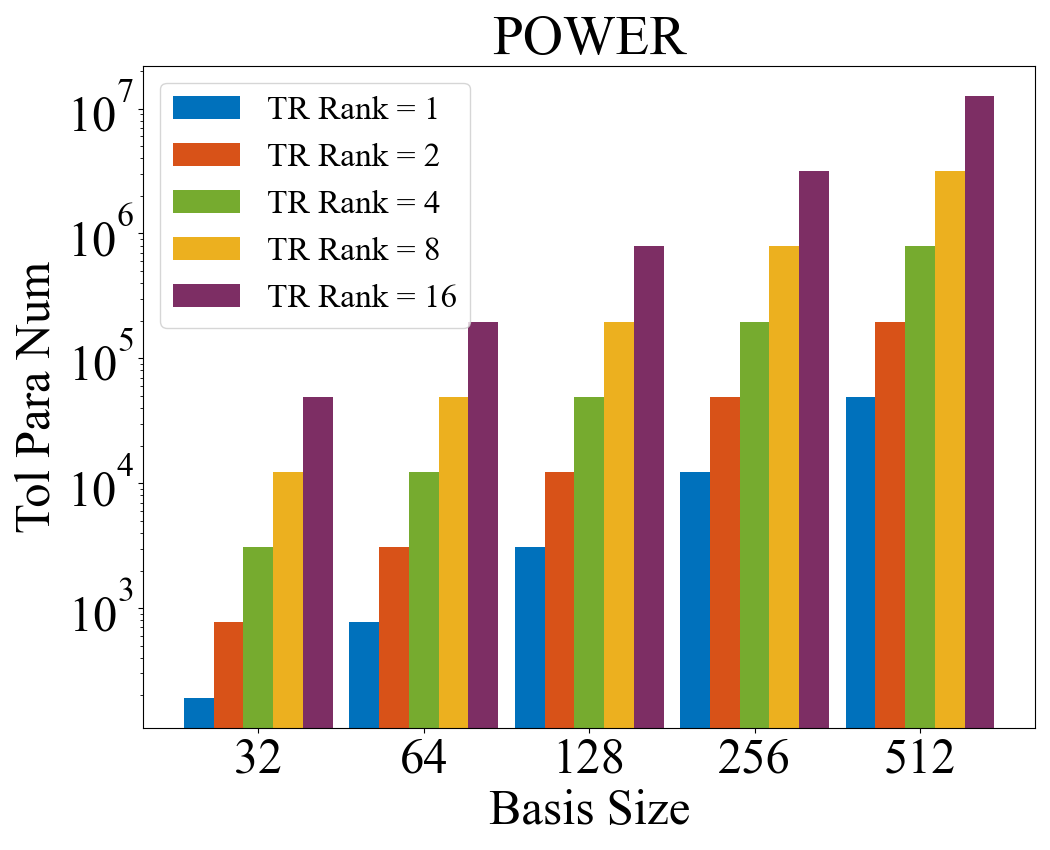}}
        \centering
            \subfigure{\includegraphics[scale=0.20]{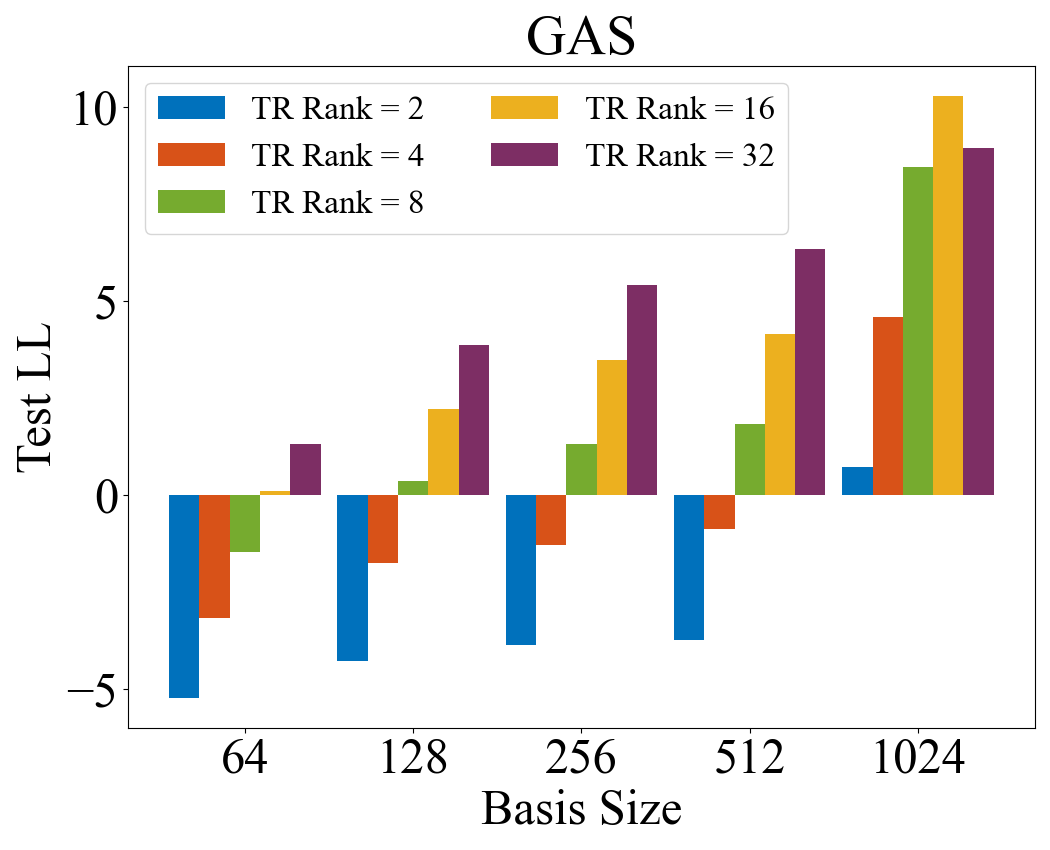}}
	    \subfigure{\includegraphics[scale=0.20]{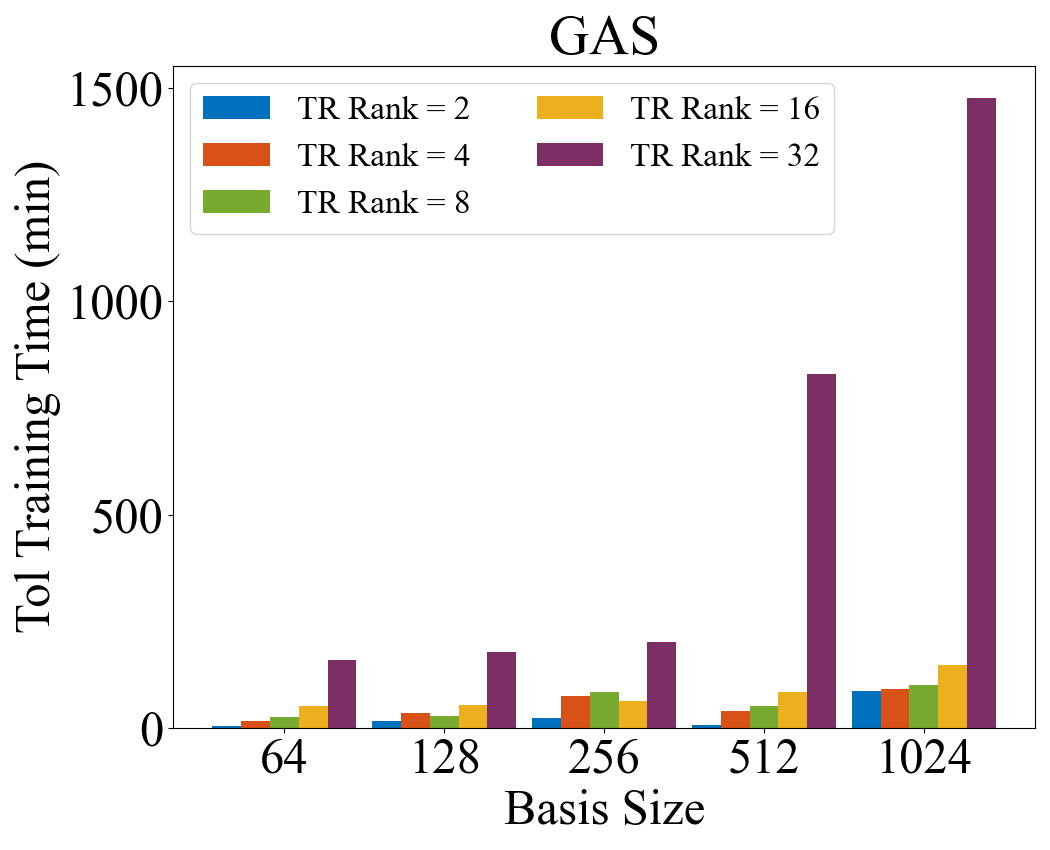}}
         \subfigure{\includegraphics[scale=0.20]{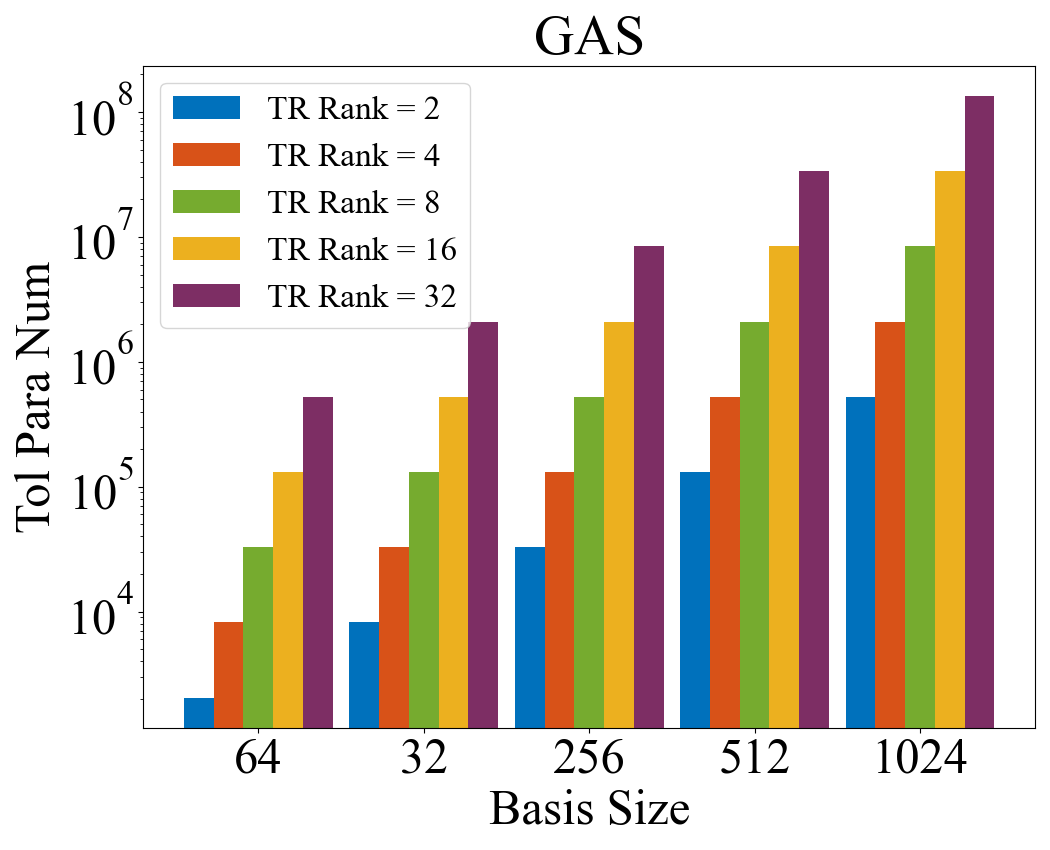}}
		\caption{ Comparison of negative log-likelihood variation, total training Time, and model parameter count with varying rank and basis size on TRDE test sets in the POWER and GAS datasets} 
        \label{power}
	\end{figure*}

In this section, we evaluate our model on four tabular UCI machine-learning datasets: POWER, GAS, HEPMASS, and MINIBOONE. These datasets were preprocessed using the method described in \cite{papamakarios2017masked}.
 \subsubsection{Comparison With Other Models}	
We compare our model against several existing advanced density estimators, including Real NVP \cite{dinh2016density}, Glow \cite{kingma2018glow}, RQ-NSF \cite{durkan2019neural}, CPFlow \cite{huang2020convex}, FFJORD \cite{grathwohl2018ffjord}, UMNN \cite{wehenkel2019unconstrained}, GF \cite{meng2020gaussianization}, and various autoregressive models such as MAF \cite{papamakarios2017masked}, MADE \cite{germain2015made}, BNAF \cite{de2020block}, and TAN \cite{oliva2018transformation}.

To ensure a thorough evaluation, we also incorporate the results of our primary competitor, TTDE, \cite{dolgov2020approximation}, for comparative analysis. Additionally, we present the baseline performance, which is derived from a Gaussian model fitted to the training data. For the experimental results, we refer to the final outcomes as reported in the aforementioned study. These consolidated results are conveniently summarized in Table~\ref{true}.
 
In POWER and GAS datasets, TERM consistently demonstrates superior performance compared to other existing normalizing flow models and autoregressive models. However, in the case of the HEPMASS and MINIBOONE datasets, where the training set size is limited, we observed notable overfitting issues, similar to those experienced with TTDE\cite{novikov2021tensor}. Despite this challenge, thanks to the more balanced structure of TRDE, our model exhibits improved performance compared to TTDE on these two datasets. Moreover, we observe that as the number of mixture components increases in these two datasets, TERM tends to accentuate overfitting. Nevertheless, when the training sample of the training set is sufficient, our method showcases remarkable capabilities in probabilistic modeling.

 \subsubsection{Parameter Impact and Trade-offs in TRDE}	
In the following analysis, we delve into the performance impact of each parameter within our model, focusing on the POWER and GAS datasets where we achieved outstanding results. Guided by TTDE\cite{dolgov2020approximation} as our benchmark with its optimal parameters, we conducted experiments by varying the basis size and TR rank of TRDE. The results are presented in Fig.~\ref{power}.

In the POWER dataset, it becomes evident that, when holding the TR rank constant, the growth in log-likelihood of the test set increases as basissize increases. Remarkably, we observe a direct proportionality between the model's total parameters and the basis size. Maintaining a constant basis size while increasing the rank leads to improved log-likelihood performance. However, due to the model's total parameters being directly proportional to the square of the TR rank, the model's size experiences exponential growth. This exponential expansion in model size is not without its trade-offs, as it results in significantly increased training time.

In the GAS dataset, we notice a clear trend: as the basis size and TR rank are increased, there is a corresponding enhancement in the model's performance. This observation is in line with our initial expectations. However, when the basis size reaches 1024 and the TR rank is set at 32, the model's training begins to show significant instability.  This occurs probably because when the number of model parameters becomes excessively large; the limited sample size is insufficient to maintain stable training and results in gradient oscillation. In addition, despite the powerful expressivity of TR, its loop structure makes the corresponding approximation problems face numerical instability and performance degradation \cite{phan2022train}, especially with larger ranks. 

 \subsubsection{Impact of TERM Components Number}	

   \begin{figure}[h!]
        \centering
		\subfigure{\includegraphics[scale=0.235]{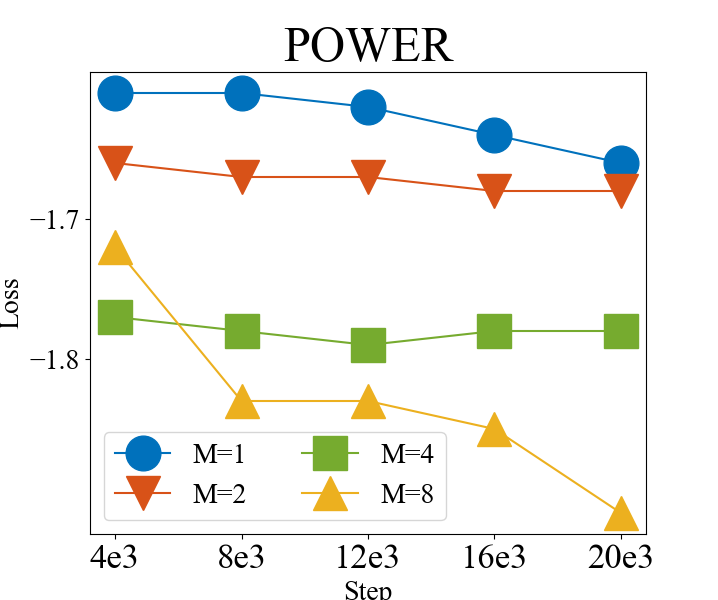}}
	    \subfigure{\includegraphics[scale=0.235]{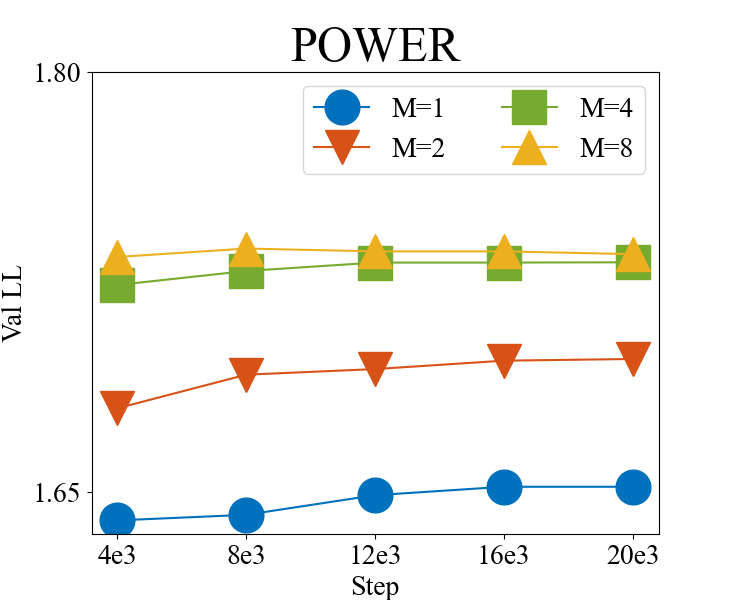}}
     \centering
		\subfigure{\includegraphics[scale=0.22]{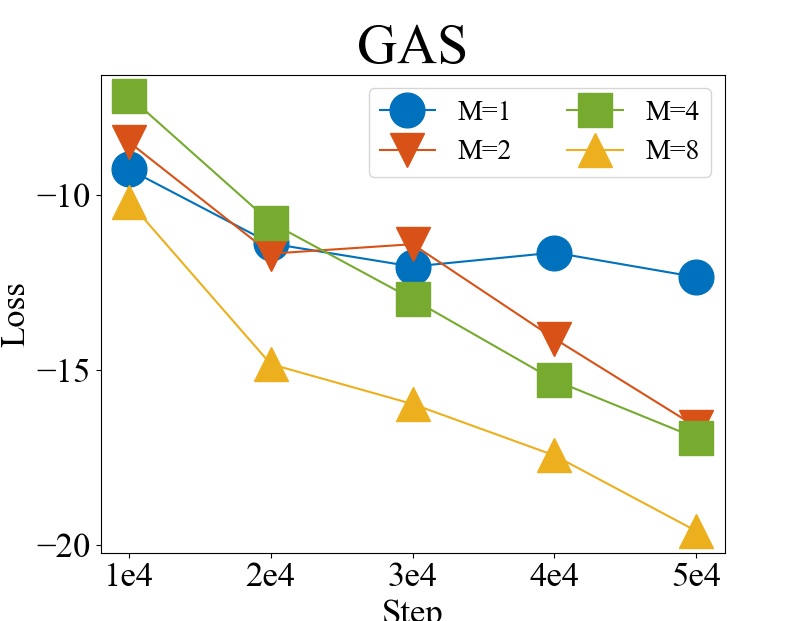}}
            \subfigure{\includegraphics[scale=0.23]{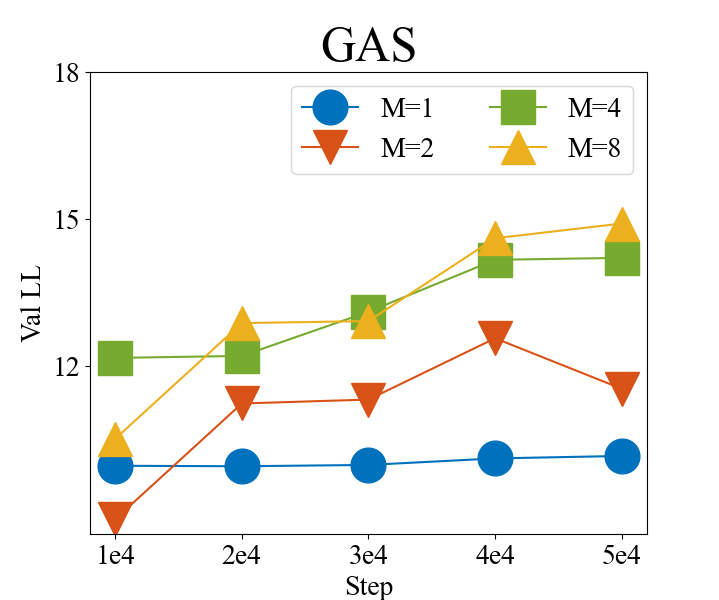}}
		\caption{ Variation of negative log-likelihoods (LL) with different components in TERM on training and validation sets during training iterations.} 
        \label{pg}
\end{figure}
To evaluate the efficacy of our proposed TERM model, we conducted an assessment on the POWER and GAS datasets, specifically examining the impact of varying the number of components. We set the TERM components to 1, 2, 4, and 8, considering the previously discussed trade-off between training time and the number of model parameters.

For the POWER dataset, TERM employ TRDE components with a basis size of 128 and a TR rank of 16. In contrast, for the GAS dataset, we use TRDE components with a basis size of 1024 and a TR rank of 16. The outcomes of these configurations are depicted in Fig.~\ref{pg}.

In our findings for the POWER dataset, the difference in performance between models using 4 and 8 components is marginal. Notably, the model with 8 components shows signs of overfitting, indicating that a 4-component TERM model is sufficient to effectively approximate the original data distribution and adequately represent the expressive space of the dataset.

In contrast, the GAS dataset displays an increase in test set likelihood with a higher number of TERM components. However, this improvement comes at the cost of increased computational and storage overhead. This observation aligns with our hypothesis that a greater number of components can better capture the underlying data structure.
\subsubsection{Comparative Analysis of Model Performance and Efficiency.}	
\begin{figure}[h!]
        \centering
		\subfigure{\includegraphics[scale=0.23]{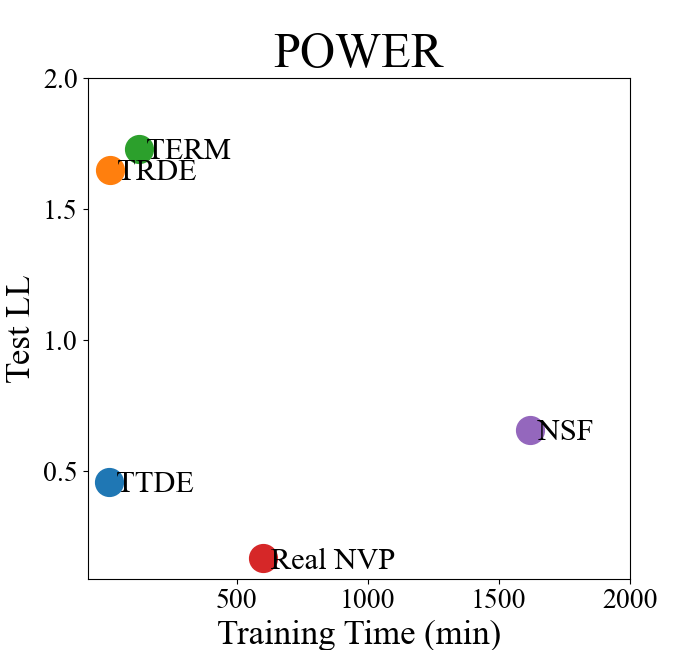}}
     \centering
		\subfigure{\includegraphics[scale=0.23]{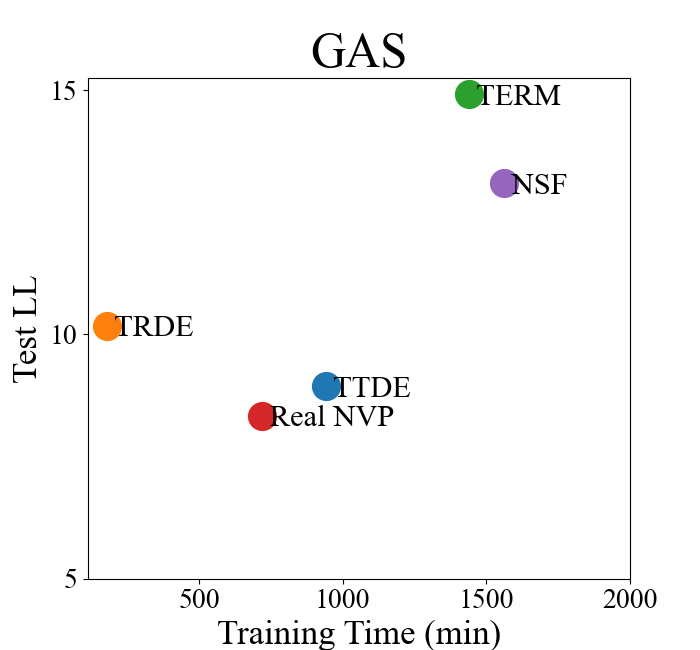}}
		\caption{ Comparative efficiency of different  methods in POWER and GAS dataset.} 
        \label{time}
\end{figure}

  In our final analysis, we compare the experimental duration and effectiveness of TRDE and TERM with the most sophisticated normalizing flow model NSF, with Real NVP and TTDE as benchmarks. 
  As depicted in Fig.~\ref{time}, our model significantly outperforms other models on the power dataset, with relatively minimal training duration and the highest log-likelihood score. For the Gas dataset, TRDE exhibits the shortest training duration and a higher log-likelihood on the test set compared to both TTDE and Real NVP. While TERM's training duration is longer than TTDE and Real NVP, it still outperforms the leading normalizing flow model, NSF, in terms of both reduced training time and enhanced test set log-likelihood.

\section{Conclusion}
In this paper, we introduce TRDE, a novel density estimator based on the tensor ring structure, enabling exact sampling, precise marginal density, and derivative calculations. Furthermore, we introduce a mixture model TERM that ensembles various permutation structures and adaptively learns the weight of each basis learner. Numerical experiments demonstrate the excellent performance of our proposed method.


\bibliographystyle{ieeetr}

\bibliography{term}

\end{document}